\documentclass[10pt,twocolumn,letterpaper]{article}

\usepackage[pagenumbers]{wacv} %

\definecolor{wacvblue}{rgb}{0.21,0.49,0.74}
\usepackage[pagebackref,breaklinks,colorlinks,allcolors=wacvblue]{hyperref}
\usepackage{booktabs}
\usepackage{siunitx}
\usepackage{caption}
\usepackage{xcolor}
\usepackage{multirow}
\usepackage{booktabs}
\usepackage{dsfont}

\def\wacvPaperID{*****} %
\def\confName{WACV}
\def\confYear{2026}

\title{Towards Consistent Long-Term Pose Generation}

\author{
Yayuan Li \quad Filippos Bellos \quad Jason J. Corso \\
University of Michigan\\
{\tt\small yayuanli@umich.edu}
}

\begin{document}
\maketitle

\begin{abstract}

Current approaches to pose generation rely heavily on intermediate representations, either through two-stage pipelines with quantization or autoregressive models that accumulate errors during inference. This fundamental limitation leads to degraded performance, particularly in long-term pose generation where maintaining temporal coherence is crucial. We propose a novel one-stage architecture that directly generates poses in continuous coordinate space from minimal context - a single RGB image and text description - while maintaining consistent distributions between training and inference. Our key innovation is eliminating the need for intermediate representations or token-based generation by operating directly on pose coordinates through a relative movement prediction mechanism that preserves spatial relationships, and a unified placeholder token approach that enables single-forward generation with identical behavior during training and inference. Through extensive experiments on Penn Action and First-Person Hand Action Benchmark (F-PHAB) datasets, we demonstrate that our approach significantly outperforms existing quantization-based and autoregressive methods, especially in long-term generation scenarios.

\end{abstract}
    
\section{Introduction}

\begin{figure}
  \centering
  \includegraphics[width=1.18\linewidth]{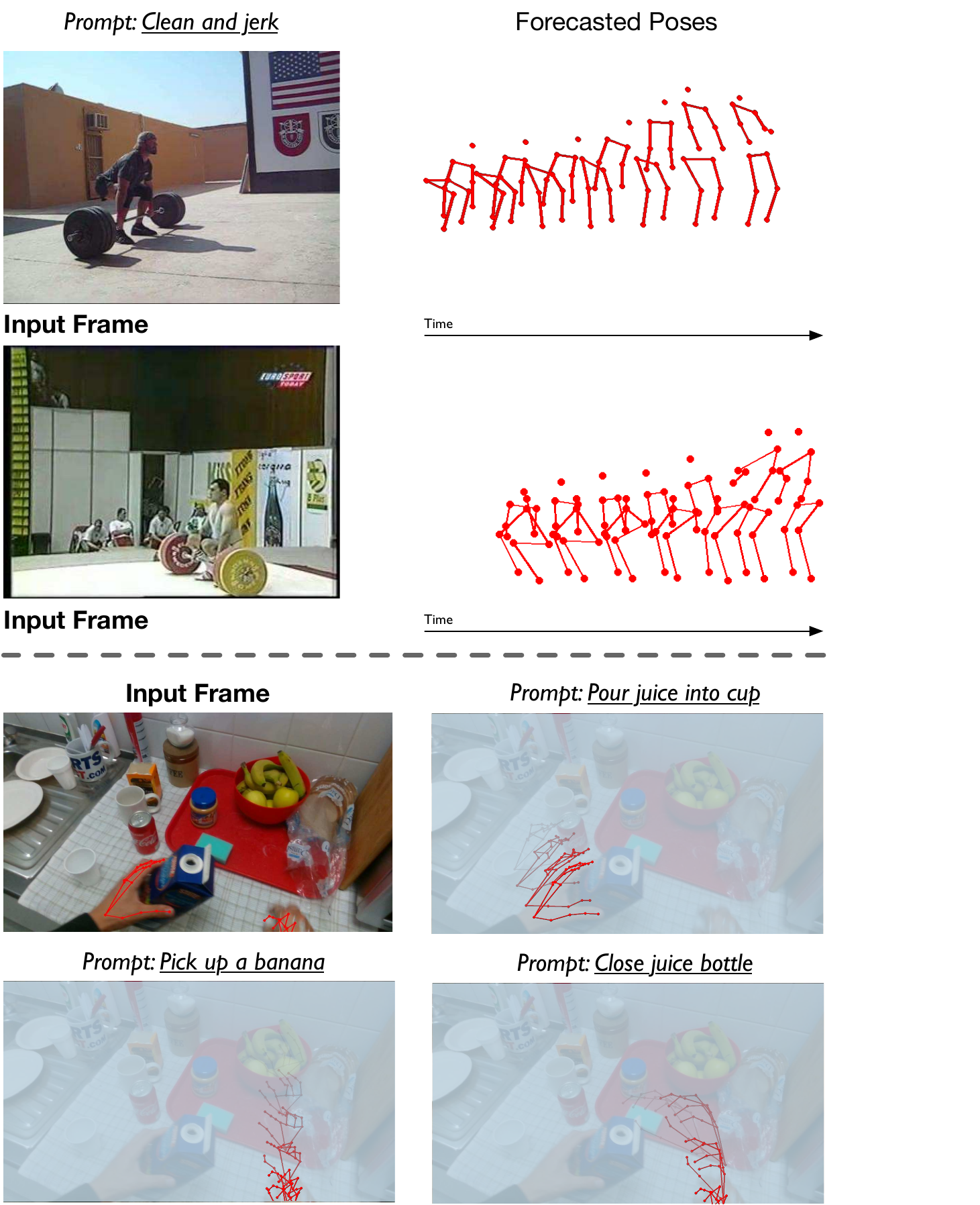}
    \caption{Examples of pose generation from a single RGB image and text description. }

  \label{fig:intro}
\end{figure}

Human pose generation has emerged as a fundamental problem in computer vision, with applications spanning animation synthesis, action understanding, and motion prediction~\cite{10.1145/3394171.3413635,chiu2019action, petrovich2021action}. Recent work has explored various approaches to control this generation process using different modalities: from textual descriptions~\cite{Ahuja2019Language2PoseNL,guo2022tm2t}, to audio signals~\cite{Lee2019DancingTM,Gong_2023_ICCV}, to scene context~\cite{wang2021synthesizing, Cao2020LongtermHM}.

Creating semantically meaningful and contextually appropriate poses remains challenging, particularly due to architectural limitations in existing approaches. These approaches typically fall into two restrictive paradigms. First, they rely on autoregressive models that generate poses frame-by-frame which injects a distribution shift between training and inference due to their nature~\cite{bachmann2024pitfallsnexttokenprediction}.
This distribution shift then leads to degraded long-term performance due to accumulated performance \cite{Dziri2023FaithAF}, as we show later in this paper. %
Second, they are two-stage approaches that first convert continuous pose coordinates into discrete tokens, latent codes through VAEs ~\cite{petrovich2021action,ma2024contact} or quantization before generation \cite{guo2022tm2t}, introducing information loss and computational overhead.

\begin{figure}
    \centering
    \includegraphics[width=1\linewidth]{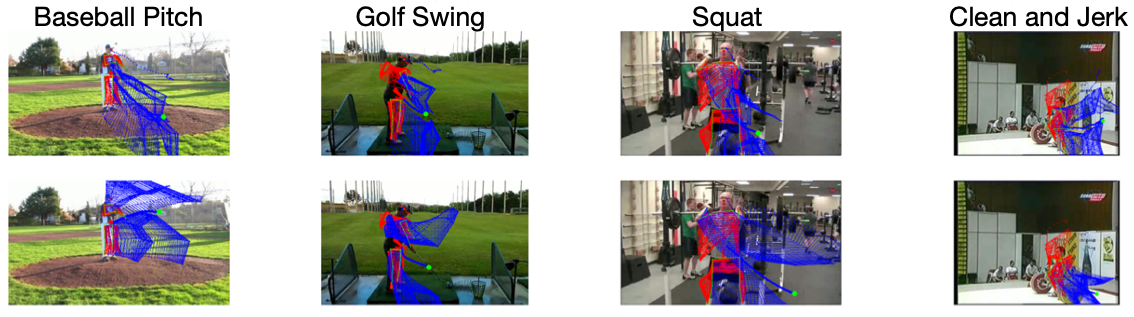}
    \caption{Examples about the issue of long term forecasting from existing methods. Red poses are the ground truth and the blues are the prediction. The error accumulates since the model are trained auto-regressively. The first row is using LSTM and the second row is using Transformer as backbone. }
    \label{fig:drift}
\end{figure}

These approaches show significant degradation when generating longer sequences, as both quantization errors and distribution shifts compound over time (as demonstrated in Figure \ref{fig:drift}). This degradation affects many downstream applications (e.g., in task guidance where long-term semantic coherence is crucial \cite{spencer2022what, damen2024genhowto}).
Additionally, most of these methods require complex inputs like 3D scene information \cite{wang2022humanise, wang2024move}, assuming the availability of such detailed data, which limits their practicality in broad real-world applications.

To address these fundamental limitations in pose generation, we introduce two key novelties within our approach:
\begin{enumerate}
    \item A unified prediction mechanism that ensures consistent distributions between training and inference, enabling reliable long-term generation.
    \item A one-stage pose generation architecture that directly operates in continuous coordinate space from minimal input---a single RGB image and text description---preserving both spatial fidelity and semantic alignment, without relying on scarce 3D detailed scene information.
    
\end{enumerate}

We also explore how language guidance can provide semantic control over the generated motions. Natural language offers an intuitive and flexible way to specify desired movements. We leverage short and concise natural language descriptions rather than the detailed movement specifications required by prior work~\cite{feng2024chatpose,lai2023lego}. This enables effective control without requiring complex movement specifications or detailed scene understanding. This combination of robust long-term generation with language control facilitates applications from animation synthesis to motion planning and task-guidance.

We evaluate the effectiveness of our method on Penn Action \cite{zhang2013actemes} and First-Person Hand Action Benchmark (F-PHAB)~\cite{FirstPersonAction_CVPR2018} datasets across different pose targets (human body and hand), viewpoints and domains. With four metrics measuring performance at different granularities, we benchmark our model against five strong baselines. Our approach consistently outperforms these baselines, achieving significant gains in both short-term and long-term pose generation. Notably, our method excels in challenging scenarios involving large motions and complex temporal dynamics. Through extensive ablation studies and qualitative results, we demonstrate that the integration of visual and textual context, along with our architecture design choices, are crucial for achieving superior performance.

\section{Related Work}

\paragraph{Pose Generation Methods}
Early approaches to pose generation (sometimes called ``forecasting") focused on predicting sequences from pose history alone, primarily handling short-term predictions of joint angles~\cite{fragkiadaki2015recurrent}. Subsequent work expanded to long-term predictions~\cite{wang2021synthesizing, huang2023diffusion}, tackling more complex motions like walking, running, and jumping. These methods, however, often rely on intermediate representations or discrete action categories, limiting their generalization capabilities.

Recent methods have explored various conditioning signals to guide pose generation. Action-conditioned approaches~\cite{chen2023humanmac, wang2021synthesizing} demonstrate success in specific categories but struggle with fine-grained control. Vision-guided methods~\cite{chao2017forecasting, fujita2023future} leverage image context but face challenges in long-term coherence. While these approaches show promise, they typically employ two-stage architectures or require conversion to intermediate representations, introducing information loss and computational overhead.

\paragraph{Language-Guided Motion Generation}
Language guidance offers rich semantic control over generated motions. Early statistical models~\cite{Takano2012BigrambasedNL, Takano2015StatisticalMC} used simple bigram representations, while recent approaches leverage deep learning architectures~\cite{petrovich2021action, tevet2023human}. Notable works like~\cite{ma2024contact} employ VQVAE and Transformer architectures, but require quantization of the continuous pose space. Others~\cite{guo2022tm2t, guo2024momask} explore text-to-motion generation but rely on complex pipelines with multiple stages.

The integration of both visual and linguistic context remains particularly challenging. While recent work~\cite{wang2024move} has attempted to combine these modalities, they often require detailed 3D scene information or rely on intermediate representations, limiting their practical applicability. In contrast, our approach operates directly in continuous coordinate space with minimal information as input.

\paragraph{Training-Inference Consistency in Generation}
Recent studies have identified fundamental limitations in traditional next-token prediction approaches~\cite{bachmann2024pitfallsnexttokenprediction, li2024autoregressive}, particularly for continuous data generation. While some works propose quantization as a solution~\cite{guo2022tm2t}, this introduces additional complexity and potential information loss. Our work directly addresses these limitations through a unified prediction mechanism that maintains consistent distributions during both training and inference.

\section{Approach}

\begin{figure*}
    \centering
    \includegraphics[width=1\linewidth]{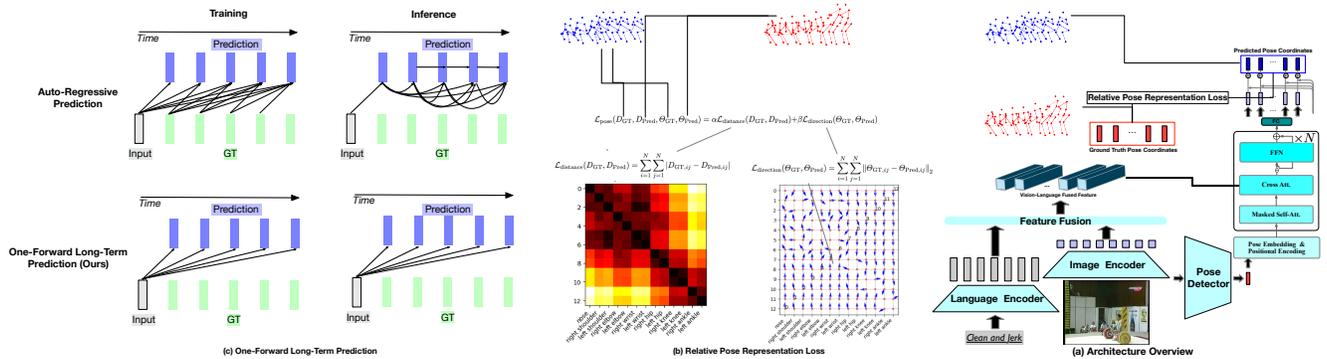}
    \caption{Overview of our proposed method. Given a single RGB image \( I \) and a natural language action description \( L \), our model extracts vision-language fused features using a multimodal encoder. These features, along with the initial pose \( P_0 \), are fed into a Transformer decoder, which predicts a sequence of future poses \( \hat{P}_{1...T} \). Our method employs cross-attention to capture the interaction between the visual and textual inputs, ensuring that the forecasted poses align with the provided context.}
\label{fig:method}
\end{figure*}

\subsection{Problem Statement}

Given a natural language description and a visual scene, our goal is to generate a sequence of future poses that accurately represents the described motion while maintaining visual context. 

Formally, given a 2D RGB image $I \in \mathbb{Z}^{H \times W \times 3}$ representing the initial scene and a language description $L = [w_1, w_2, ..., w_M]$ comprising $M$ tokenized words, our objective is to generate a pose sequence $\mathbf{P} = \{P_i\}_{i=1}^k$ that is both semantically consistent with the language description and visually coherent with the scene, where each sequence consists of $k$ future frames. Unlike previous works that require complex 3D scene data~\cite{wang2024move}, our method relies only on a single RGB image 
We parameterize each pose $P_i \in \mathds{R}^{2N}$ using 2D keypoint coordinates, where $N$ represents the number of keypoints (13 for human body poses capturing joints like head, shoulders, and elbows, or 21 for hand poses representing joints like wrist and fingers). In contrast to previous approaches that require intermediate representations, we operate directly in this continuous coordinate space.
\subsection{Method}
\label{model_structure}
We propose a novel one-stage architecture for generating poses directly in continuous coordinate space from visual and textual inputs. As illustrated in Fig.~\ref{fig:method}, our model fuses information from a single RGB image and language description through three key components. A vision-language encoder first processes the image $I$ through an Image Encoder $f_I$ to extract features $F_I \in \mathds{R}^{N_I \times d_I}$, while simultaneously processing the language description $L$ through a Multimodal Feature Fusion module $f_M$ to generate fused image-text features $F_M \in \mathds{R}^{N_M \times d_M}$. These features feed into our pose forecasting module, which employs a placeholder token mechanism to predict future poses relative to the detected initial pose, enabling unified training and inference in a single forward pass.

\paragraph{Training-Inference Alignment for Consistent Pose Generation} Predicting the next token has shown strong performance across various tasks. However, recent studies have identified limitations due to the distribution shift between training and inference. In classification tasks, this shift may limit the model's ability to learn long-term dependencies, although large training datasets can somewhat mitigate the issue~\cite{brown2020language}. However, we found that for regression tasks like pose forecasting, this issue is more pronounced, leading to significant degradation in long-term predictions. Although recent approaches~\cite{guo2022tm2t} attempt to address this by quantizing poses before next-token prediction, they rely on two-stage training, with final performance heavily dependent on the quality of the quantization and reconstruction steps.

In contrast, we propose a one-stage approach that directly forecasts continuous pose coordinates (as shown in ~\cref{fig:method}-(c)), maintaining strong performance~\cref{tab:table.baseline} compared to previous two-stage methods. Our model predicts multiple future tokens simultaneously from a single input during both training and inference, using a placeholder token [PRD] and non-masked self-attention for efficient decoding.

We observe that it is the mismatch between input distributions during training and inference that essentially leads to a drop in accuracy for pose generation. To address this, we introduce a method that aligns the input distributions, significantly improving long-term prediction accuracy. Our Transformer Decoder processes the initial pose \( P_0 \), which is extracted from $I$ by a pre-trained pose detector~\cite{lugaresi2019mediapipe},  combined with fused visual-linguistic features ${F_{M}}$, to generate future poses:
\begin{equation}
\hat{P} = \text{Decoder}(P_0, {F_{M}}) \in \mathds{R}^{T \times 2N}
\end{equation}
The predicted poses \( \hat{P} \) are then compared against the ground truth \( P_{1...T} \in \mathds{R}^{T \times 2N} \).

In the previous next-token prediction (NTP) approach, the decoder input during training is structured as:
\begin{equation}
X^{NTP} = \begin{pmatrix}
x^{0}_{1} & y^{0}_{1} & \cdots & x^{0}_{N} & y^{0}_{N} \\
x^{1}_{1} & y^{1}_{1} & \cdots & x^{1}_{N} & y^{1}_{N} \\
\vdots & \vdots & \ddots & \vdots & \vdots \\
x^{T-1}_{1} & y^{T-1}_{1} & \cdots & x^{T-1}_{N} & y^{T-1}_{N} 
\end{pmatrix} 
\end{equation} where $X^{NTP} \in \mathds{R}^{T \times 2N}$. 

The Masked Self-Attention mechanism ensures that each prediction \( \hat{P_i} \) (\(i = 1, 2, ..., T\)) relies on the preceding poses \( P_{0 ... i-1} \). However, during inference, the reliance on predicted poses introduces errors that accumulate over time, degrading the performance of long-term forecasting.

To resolve this, we propose a new input structure that ensures consistency between training and inference. This structure uses placeholder tokens for future timestamps, allowing the model to predict all future poses in a single forward pass:
\begin{equation}
X^{ours} = \begin{pmatrix}
x^{0}_{1} & y^{0}_{1} & \cdots & x^{0}_{N} & y^{0}_{N} \\
[\text{PRD}]_{1} & [\text{PRD}]_{2} & \cdots & [\text{PRD}]_{2N} \\
\vdots & \vdots & \ddots & \vdots & \vdots \\
[\text{PRD}]_{1} & [\text{PRD}]_{2} & \cdots & [\text{PRD}]_{2N}
\end{pmatrix}
\end{equation} where $X^{ours} \in \mathds{R}^{T \times 2N}$.

Here, the placeholder token \([\text{PRD}] \in \mathds{R}^{2N}\) contains no information but marks the positions to be predicted. This approach eliminates the need to rely on ground-truth poses during forwarding (for both training and inference), avoiding error accumulation in the decoder during inference. The positional encoding remains the sole distinguishable information, guiding the model to produce distinct predictions for each timestamp. This design allows a single stage approach for pose generation in coordinate space, keeping the inference performance. Furthermore, the inference efficiency is improved since only one forward is needed to produce all poses in batch.

\paragraph{Generating Temporally Relative Pose Movement} Our model generates pose sequences by predicting relative movements from an initial pose. Rather than predicting absolute coordinates, we leverage relative displacement prediction. Consider a scenario where a head joint needs to move downward: instead of directly predicting its final position ($x=0.7, y=0.9$) in global coordinates, our model first detects its current position ($x=0.75, y=0.8$) in the input image, then predicts the required displacement ($\Delta{x}=-0.05$, $\Delta{y}=0.1$) to reach the target. This approach provides two key advantages: it incorporates spatial context from the initial pose (e.g., right side or left side of the image) and directly models movement as displacement rather than absolute positions. 

The effectiveness of this design is demonstrated in our ablation study.

\paragraph{Vision-Language Feature Encoding} The multimodal features are fused by the vision-language encoder. The Image Encoder $f_I$ processes the visual input $I$ to produce image features $F_I \in \mathds{R}^{N_I \times d_I}$. For language representation, we rely on succinct natural language terms (e.g., \emph{``swing golf''}), contrasting with prior work~\cite{feng2024chatpose,lai2023lego} that require detailed, less practical text descriptions (e.g., \emph{``The upper body turns to the right-hand side, hips twist left-hand side, and arms move the club down in a curved path to hit the ball''}). The Multimodal Feature Fusion module $f_M$ integrates these compact descriptions with the visual features to generate fused features $F_M \in \mathds{R}^{N_M \times d_M}$. In practice, we employ BLIP's~\cite{li2022blip} Unimodal Encoder module for $f_I$ and its Image-grounded text encoder for $f_M$, keeping the pretrained weights frozen.

\subsection{Relative Pose Representation Loss}
To accurately capture the intrinsic spatial relationships between joints and reduce redundancy in global coordinates, we design a loss function that combines relative distances and directions between joints~\cite{cao2017realtime} with a Mean Squared Error (MSE) component. This formulation enhances pose prediction accuracy by emphasizing the spatial coherence of joint movements, ensuring that the model learns both the relative positioning and the overall structure of the pose.

\paragraph{Distance Representation} We represent the pairwise Euclidean distances between adjacent joints in a distance matrix \( D \in \mathds{R}^{N \times N} \):
\begin{equation}
D_{ij} = \sqrt{(x_i - x_j)^2 + (y_i - y_j)^2}
\end{equation}
where \( (i, j) \) are adjacent joints.

\paragraph{Direction Representation} The direction matrix \( \Theta \) encodes the unit direction vectors between adjacent joints:
\begin{equation}
\Theta_{ij} = \left( \frac{x_j - x_i}{D_{ij}}, \frac{y_j - y_i}{D_{ij}} \right)
\end{equation}

\paragraph{Loss for a Single Pose:} The total pose loss is a weighted sum of the distance and direction losses:

\paragraph{Distance Loss}
\begin{equation}
\mathcal{L}_{\text{distance}}(D_{\text{GT}}, D_{\text{Pred}}) = \sum_{i=1}^{N} \sum_{j=1}^{N} \left| D_{\text{GT},ij} - D_{\text{Pred},ij} \right|
\end{equation}

\paragraph{Direction Loss}
\begin{equation}
\mathcal{L}_{\text{direction}}(\Theta_{\text{GT}}, \Theta_{\text{Pred}}) = \sum_{i=1}^{N} \sum_{j=1}^{N} \left\| \Theta_{\text{GT},ij} - \Theta_{\text{Pred},ij} \right\|_2
\end{equation}

\paragraph{Total Pose Loss}
\begin{equation}
\mathcal{L}_{\text{pose}} = \alpha \mathcal{L}_{\text{distance}} + \beta \mathcal{L}_{\text{direction}}
\end{equation}
where \( \alpha \) and \( \beta \) control the relative contributions of each loss component.

\paragraph{Loss for a Sequence of Poses} Given a sequence of ground truth poses \( P_{1...k} \in \mathds{R}^{k \times 2N} \) and predicted poses \( \hat{P}_{1...k} \in \mathds{R}^{k \times 2N} \), the total sequence loss is computed as the mean over all timestamps:
\begin{equation}
\mathcal{L}_{\text{seq}} = \frac{1}{k} \sum_{i=1}^{k} \mathcal{L}_{\text{pose}}(D_{\text{GT}}, D_{\text{Pred}}, \Theta_{\text{GT}}, \Theta_{\text{Pred}})
\end{equation}

\paragraph{Batch Loss} When optimizing over multiple samples, the batch loss is the mean of sequence losses:
\begin{equation}
\mathcal{L}_{\text{batch}} = \frac{1}{B} \sum_{i=1}^{B} \mathcal{L}_{\text{seq}, i}
\end{equation}
where \( B \) is the batch size.

\paragraph{Final Loss} The final loss function used to train the model is a combination of the relative pose loss and a standard MSE loss:
\begin{equation}
\mathcal{L} = \mathcal{L}_{\text{rel}}(\alpha, \beta) + \theta \mathcal{L}_{\text{batch}, mse}
\end{equation}
where \( \theta \) controls the contribution of the MSE term.

\section{Experiments}
In this section, we demonstrate the performance of our method by quantitative and qualitative results. We introduce our experiment settings including dataset, evaluation metrics and baselines. We compare our method with the strong baseline methods and with previous SOTA pose generation methods. We investigate the performance at finer grain in terms of forecasting steps and hardness. Also, we do ablation study to demonstrate the effeciency of our designs.

\subsection{Dataset}
\begin{table*}[t]
  \centering 
  \setlength{\tabcolsep}{5pt}
  \begin{tabular}{l *{8}{S[table-format=1.2]}}
  \toprule
  & \multicolumn{4}{c}{Penn Action} & \multicolumn{4}{c}{F-PHAB} \\
  \cmidrule(lr){2-5} \cmidrule(lr){6-9}
  {Method} & {ADE $\downarrow$} & {FDE $\downarrow$} & {PCK $\uparrow$} & {RMSE $\downarrow$} 
  & {ADE $\downarrow$} & {FDE $\downarrow$} & {PCK $\uparrow$} & {RMSE $\downarrow$} \\
  \midrule
  $\text{NN}_{\text{P}}$ & {0.0901} & {0.1050} & {0.6663} & {0.0568} & {0.1684} & {0.1540} & {0.3769} & {0.1091} \\
  $\text{NN}_{\text{VL}}$ & {0.2421} & {0.2461} & {0.2997} & {0.1566} & {0.2583} & {0.2588} & {0.2793} & {0.2143} \\
  LSTM~\cite{10.1162/neco.1997.9.8.1735} & {0.1635} & {0.2622} & {0.3820} & {0.1061} & {0.1938} & {0.1938} & {0.3018} & {0.1358} \\
  Naive Transformer~\cite{vaswani2017attention} & {0.1726} & {0.2303} & {0.3435} & {0.1110} & {0.1922} & {0.2031} & {0.3001} & {0.1459} \\
  Quantization + Transformer~\cite{guo2022tm2t} & {0.2549} & {0.2478} & {0.1798} & {0.1664} & {0.2431} & {0.2387} & {0.2077} & {0.1601} \\
  Ours & {\textbf{0.0578}} & {\textbf{0.0766}} & {\textbf{0.8179}} & {\textbf{0.0350}} & {\textbf{0.0967}} & {\textbf{0.0855}} & {\textbf{0.7645}} & {\textbf{0.0683}} \\
  \bottomrule
  \end{tabular}
  \caption{Comparison of our method with baselines on Penn Action and F-PHAB datasets using ADE, FDE, PCK, and RMSE metrics. Our method consistently outperforms all baselines, particularly in long-term generation.}
  \label{tab:table.baseline}
\end{table*}

\begin{figure*}
    \centering
    \includegraphics[width=1\linewidth]{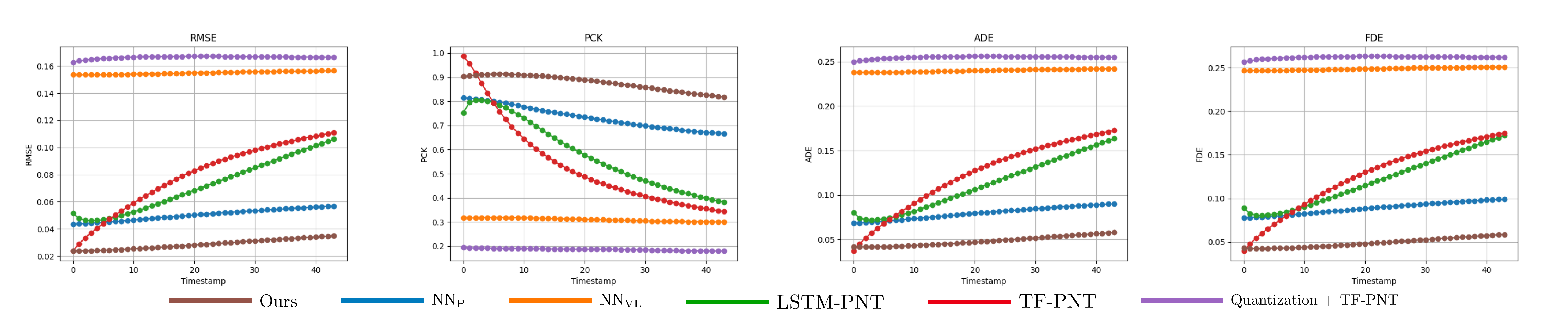}
    \caption{Performance across generation timestamps. Our method consistently outperforms baseline methods across most of the timestamps.}
    \label{fig:bytime_byhard}
\end{figure*}

We evaluate our method on diverse datasets to demonstrate its generalization across different targets (human body and hand) and scenarios. All datasets are video-based, where each clip is annotated with natural language action descriptions. For datasets without pose annotations, we use Mediapipe to generate pseudo-annotations. We split 90\% of the data for training and 10\% for testing, using forecasting horizon of 45 timestamps for training and evaluating all forecasting steps from 1 to 45 frames.

We use two datasets: Penn Action~\cite{zhang2013actemes} and First-Person Hand Action Benchmark (F-PHAB)~\cite{FirstPersonAction_CVPR2018}. They cover two targets (human body and human hand), multiple domains (e.g., cooking, sports) and viewpoints (egocentric and exocentric).

These datasets allow us to explore various use cases, from fine-grained hand movements to larger human body motions, across different domains and viewpoints.

\subsection{Implementation details}
In our implementation, we use a frozen BLIP to
fuse the multimodal features. The unimodal encoder used is ViT-g/14 and the text encoder is BERT. The normalization factor
$\sigma$ is set to 0.8. The Transformer models are constructed using
the native PyTorch implementation. Our model
undergoes training to convergence using the AdamW optimizer
with a fixed learning rate of $10^{-4}$. The training of our method is conducted on 1 NVIDIA H100 GPUs, assigning a batch size
of 64 per GPU.

\subsection{Metrics}

We evaluate performance using four metrics that capture different aspects of pose generation:

\paragraph{Root Mean Square Error (RMSE)} Measures the average distance between generated and annotated keypoints, providing a normalized error based on the size of the input image. Formally:
\begin{equation}
\text{RMSE}(\hat{Y}, Y) = \sqrt{\frac{1}{T \times N} \sum_{t=1}^{T} \sum_{n=1}^{N} (\hat{Y}_{tn} - Y_{tn})^2}
\end{equation}

\paragraph{Percentage of Correct Keypoints (PCK)} The percentage of keypoints whose generated position is within a certain threshold (\(\delta\)) of the annotated position. We set \(\delta=0.05\) for human targets and \(\delta=0.15\) for hands:
\begin{equation}
\text{PCK}(\hat{Y}, Y) = \frac{1}{T \times K} \sum_{t=1}^{T} \sum_{k=1}^{K} \mathbb{I}(||\hat{Y'}_{tk} - Y'_{tk}|| < \delta)
\end{equation}

\paragraph{Average Displacement Error (ADE)} Measures the average \(l_2\) distance between generated and ground-truth trajectories over all timestamps:
\begin{equation}
\text{ADE}(\hat{Z}, Z) = \frac{1}{T} \sum_{t=1}^{T} ||\hat{Z}_{t} - Z_{t}||
\end{equation}

\paragraph{Final Displacement Error (FDE)} Measures the \(l_2\) distance between the generated and ground-truth trajectories at the final timestamp:
\begin{equation}
\text{FDE}(\hat{Z}, Z) = ||\hat{Z}_{T} - Z_{T}||
\end{equation}
\subsection{Baselines}
\label{exp_baseline}
\begin{figure*}
  \centering
  \includegraphics[width=\textwidth]{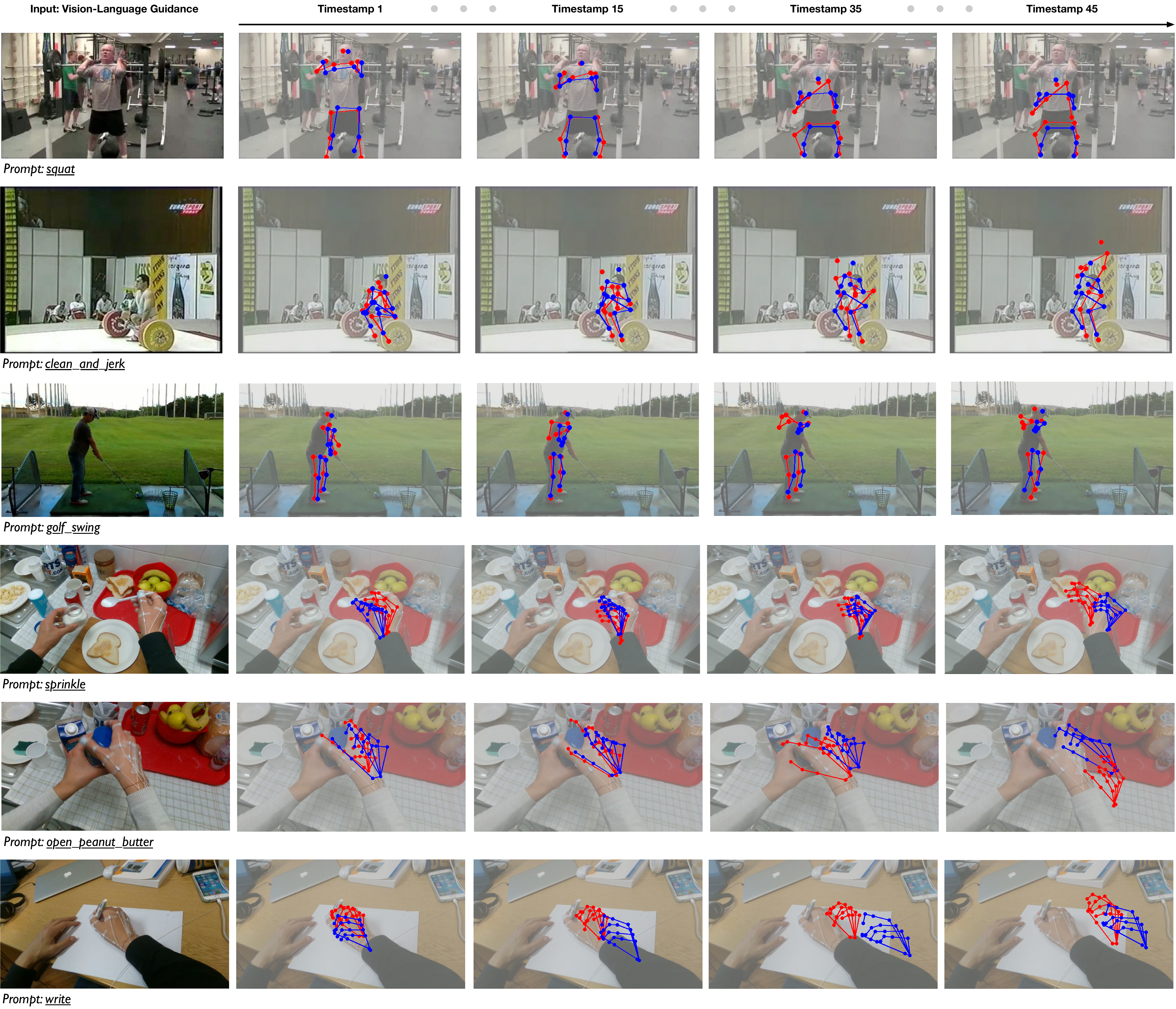}
  \caption{Qualitative results of our proposed method. Red poses are annotations, and blue poses are predictions from our method. }
  \label{fig:result.baseline}
\end{figure*}

We introduce five baselines inspired by previous works to benchmark our newly proposed problem of generating instructional poses from a single RGB image and text. These baselines include Nearest Neighbor (with different similarity features), LSTM-based pose generation, the Original Transformer Decoder, and a two-stage generation method with Pose Quantization.

\paragraph{Nearest Neighbor (NN)} This baseline retrieves the most similar sample from the training data based on the input image and text narration, using its corresponding pose as the prediction. It measures how well existing pose sequences can serve as predictions for test samples. If this method performs well, it suggests that testing outputs are well-represented in the training set. We explore two variations:
\begin{itemize}
    \item \(\text{NN}_{\text{P}}\): Similarity is computed as the Euclidean distance between keypoint coordinates in the input and training images.
    \item \(\text{NN}_{\text{VL}}\): Similarity is based on the Euclidean distance between fused vision and language features.
\end{itemize}

\paragraph{LSTM} We use an LSTM-based model for pose generation, leveraging vision-language features as input.

\paragraph{Naive Transformer} This baseline uses the Original Transformer Decoder with a next-token prediction task to compare its performance with our proposed solution.

\paragraph{Quantization + Transformer} Inspired by state-of-the-art single-modality pose generation, this two-stage baseline first quantizes pose coordinates using VQ-VAE and then applies a Transformer Decoder. This allows us to evaluate the efficiency of the two-stage approach in our scenario. 

\subsection{Results}
\begin{table*}[t]
  \centering 
  \setlength{\tabcolsep}{5pt}
  \begin{tabular}{l *{8}{S[table-format=1.2]}}
  \toprule
  & \multicolumn{4}{c}{Penn Action} & \multicolumn{4}{c}{F-PHAB} \\
  \cmidrule(lr){2-5} \cmidrule(lr){6-9}
  {Method} & {ADE $\downarrow$} & {FDE $\downarrow$} & {PCK $\uparrow$} & {RMSE $\downarrow$} 
  & {ADE $\downarrow$} & {FDE $\downarrow$} & {PCK $\uparrow$} & {RMSE $\downarrow$} \\
  \midrule
  TF & {0.1726} & {0.2303} & {0.3435} & {0.1110} & {0.1922} & {0.2031} & {0.3001} & {0.1459} \\
  \hspace{1em} + Pose Det. & {0.1246} & {0.1548} & {0.4412} & {0.0978} & {0.1638} & {0.1702} & {0.3323} & {0.1287} \\
  \hspace{2em} + Train in Batch (Full Attn) & {0.0693} & {0.0693} & {0.7741} & {0.0431} & {0.1425} & {0.1433} & {0.3847} & {0.0908} \\
  \hspace{3em} + Causal Mask & {0.0595} & {0.0742} & {0.8204} & {0.0370} & {0.1231} & {0.1335} & {0.4210} & {0.0823} \\
  \hspace{4em} + Ours (relative loss) & {\textbf{0.0578}} & {\textbf{0.0766}} & {\textbf{0.8179}} & {\textbf{0.0350}} & {\textbf{0.0967}} & {\textbf{0.0855}} & {\textbf{0.7645}} & {\textbf{0.0683}} \\
  \bottomrule
  \end{tabular}
  \caption{Ablation study results comparing different module choices on Penn Action and F-PHAB using ADE, FDE, PCK, and RMSE.}
  \label{tab:table.ablation}
\end{table*}

We present the first results for the newly proposed problem of vision-language-guided pose generation from a single RGB image. We compare our method against five baselines using the metrics introduced earlier. Additionally, we investigate performance across different timestamps, analyze hard samples, perform an ablation study, and compare with state-of-the-art single-modality methods to demonstrate the necessity of integrating both visual and textual information.

\paragraph{Main Results} 
Table~\ref{tab:table.baseline} summarizes the performance of our method compared to five challenging baselines. Our method outperform all of the baselines. When traditional Nearest Neighbor (NN) incoporates with extracted information from advanced deep networks (e.g., pose detector or Vision-Language feature extractor), it gives surprisingly good performance ($\text{NN}_{\text{P}}$ and $\text{NN}_{\text{VL}}$). In addition, LSTM-NTP and TF-NTP demonstrate the limitation of training with next-token prediction (NTP) on two popular deep network structure --- LSTM and Transformer. NTP fails to help the model to learn motion longer than one future timestamps during training and leads to ```drifting''' behavior of pose generation during inference (Figure \ref{fig:drift}). Inspired by previous works~\cite{guo2022tm2t}, we do quantization before NTP with Transformer (Quantization + TF-NTP). However, as we argue in~\cref{model_structure}, this approach heavily relies on the quantization performance which require big amount of data to construct the discrete space in self-supervised manner. We demonstrate that although the NTP give decent performance at quantized space, reconstructing the quantized poses to actual pose coordinates becomes a performance bottleneck and leads to bad overall performance. 

On the other hand, our method, designed to generate future poses directly in coordinate space with one-forward prediction (predict multiple tokens in one forward)~\cref{model_structure}, successfully overcome the ``drifting" behavior in LSTM-NPT and TF-NPT and avoid the information loss in quantization process (Quantization + TF-NTP). This improvement is shown in both quantitative~\cref{tab:table.baseline} and qualitative results~\cref{fig:result.baseline}.

\paragraph{Performance by Timestamp and Sample Hardness} We further analyze the performance of each method across different timestamps and on hardest 10\% testing samples. 

As shown in Figure~\ref{fig:bytime_byhard}, our method consistently outperforms the baselines across most forecasting horizons especially the longer ones. Furthermore, our method maintains robust performance as forecasting horizon increases, in contract with the dramatic performance drop of LSTM-NTP and TF-NTP. This highlights the efficiency of our design of predicting multiple future timestamps in one forward.

Also, ~\cref{tab:table.hardness} shows that our method maintain the best performance in hardest 10\% of testing samples. The hardness is defined by the variance of keypoint motion. I.e., more motion in the ground truth future pose sequences indicates that it is harder to capture the dynamic.

\begin{table}[t]
  \centering 
  \small %
  \setlength{\tabcolsep}{3pt}
  \begin{tabular}{l *{4}{S[table-format=1.2]}}
  \toprule
  & \multicolumn{4}{c}{Penn Action} \\
  \cmidrule(lr){2-5} 
  {Method} & {ADE $\downarrow$} & {FDE $\downarrow$} & {PCK $\uparrow$} & {RMSE $\downarrow$} 
  \\
  \midrule
  $NN_{P}$ & {0.1118} & {0.1653} & {0.5491} & {0.0703}  \\
  $NN_{VL}$ & {0.2247} & {0.2577} & {0.1813} & {0.1446} \\
  LSTM & {0.1735} & {0.2886} & {0.3159} & {0.1121} \\
  Naive Transformer & {0.1680} & {0.2546} & {0.3265} & {0.1076}  \\
  Quantization + Transformer & {0.2467} & {0.2556} & {0.1235} & {0.1586} \\
  Ours & {\textbf{0.0922}} & {\textbf{0.1573}} & {\textbf{0.6824}} & {\textbf{0.0573}} \\
  \bottomrule
  \end{tabular}
  \caption{Results on hardest 10\% test samples. The hardness is defined by the variance in annotated pose sequence.}
  \label{tab:table.hardness}
\end{table}

\paragraph{Ablation Studies}
Table~\ref{tab:table.ablation} shows the results of our ablation studies. We compare the effects of different design choices, such as encoder selection and the use of in-batch training. Our full model, which generating all future poses by generation from the detected initial pose in one-forwarding along with the relative loss function, consistently outperforms simplified versions. These results underline the importance of our design choices for achieving optimal performance.

\paragraph{Comparison with SOTA Single-Modality Pose Generation Methods}
We compare our method with state-of-the-art (SOTA) single-modality pose generation methods to highlight the necessity of using both visual and textual information for generating instructional poses. It is important to note that this comparison is not entirely fair, as previous methods utilize stronger inputs, such as multiple frames, while we use only a single RGB image and simple text.

Table~\ref{tab:sota} shows that our method significantly outperforms SOTA text-only method TM2T~\cite{guo2022tm2t} and SOTA vison-only method~\cite{zhang2019predicting}.

\begin{table}
  \centering
  \setlength{\tabcolsep}{7.2pt}
  \begin{tabular}{l *{4}{S[table-format=1.2]}}
  \toprule
  & \multicolumn{4}{c}{Penn Action} \\
  \cmidrule(lr){2-5}
  {Method} & {ADE $\downarrow$} & {FDE $\downarrow$} & {PCK $\uparrow$} & {RMSE $\downarrow$} \\
  \midrule
  TM2T~\cite{guo2022tm2t} & {0.2684} & {0.2924} & {0.1709} & {0.2708} \\
  PHD~\cite{zhang2019predicting} & {-} & {-} & {0.7720} & {-} \\
  Ours & {\textbf{0.0169}} & {\textbf{0.0170}} & {\textbf{0.860}} & {\textbf{0.012}} \\
  \bottomrule
  \end{tabular}
  \caption{Comparison with SOTA single modality generation methods on Penn Action dataset. }
  \label{tab:sota}
\end{table}

\section{Conclusion}

In this work, we introduced a novel one-stage architecture for vision-language-guided pose forecasting, which operates directly in continuous coordinate space. By maintaining consistency between training and inference and leveraging relative movement prediction, our method achieves superior pose generation while preserving spatial fidelity. Through extensive experiments on standard pose datasets, including Penn Action and F-PHAB, our approach demonstrated state-of-the-art performance across multiple metrics, significantly outperforming SOTA baselines. Furthermore, our design excels in handling challenging scenarios, such as large motions and dynamic temporal sequences, underscoring its robustness for long-term forecasting.

{
    \small
    \bibliographystyle{ieeenat_fullname}
    \bibliography{main}
}

\end{document}